\documentclass[anonymous=false,acmsmall,screen,nonacm]{acmart}

\usepackage[utf8]{inputenc}
\usepackage{blindtext}
\usepackage{subfiles}   
\usepackage{graphicx}
\usepackage{array}
\usepackage{subfiles} 
\usepackage{amsmath,graphicx}
\usepackage{enumitem}
\usepackage{multirow}
\usepackage{balance}

\makeatletter
\def\runningfoot{\def\@runningfoot{}}
\def\firstfoot{\def\@firstfoot{}}
\let\@authorsaddresses\@empty
\makeatother

\renewcommand\footnotetextcopyrightpermission[1]{} 
\setcopyright{none}

\begin{document}

\title{The Bystander Affect Detection (BAD) Dataset \\
for Failure Detection in HRI
}

\author{Alexandra Bremers}
\affiliation{%
  \institution{Cornell Tech}
  \country{USA}}
\email{awb227@cornell.edu}

\author{Maria Teresa Parreira}
\affiliation{%
  \institution{Cornell Tech}
  \city{New York}
  \country{USA}}

\author{Xuanyu Fang}
\affiliation{%
  \institution{Cornell Tech}
  \city{New York}
  \country{USA}}

\author{Natalie Friedman}
\affiliation{%
  \institution{Cornell Tech}
  \city{New York}
  \country{USA}}

\author{Adolfo Ramirez-Aristizabal}
\affiliation{%
  \institution{Accenture Labs}
  \city{San Francisco}
  \country{USA}}

\author{Alexandria Pabst}
\affiliation{%
  \institution{Accenture Labs}
  \city{San Francisco}
  \country{USA}}

\author{Mirjana Spasojevic}
\affiliation{%
  \institution{Accenture Labs}
  \city{San Francisco}
  \country{USA}}

\author{Michael Kuniavsky}
\affiliation{%
  \institution{Accenture Labs}
  \city{San Francisco}
  \country{USA}}

\author{Wendy Ju}
\affiliation{%
  \institution{Cornell Tech}
  \city{New York}
  \country{USA}}

\thispagestyle{empty}
\pagestyle{empty}

\begin{abstract}

 For a robot to repair its own error, it must first know it has made a mistake. One way that people detect errors is from the implicit reactions from bystanders -- their confusion, smirks, or giggles clue us in that something unexpected occurred. 
 To enable robots to detect and act on bystander responses to task failures, we developed a novel method to elicit bystander responses to human and robot errors. Using 46 different stimulus videos featuring a variety of human and machine task failures, we collected a total of 2452 webcam videos of human reactions from 54 participants. To test the viability of the collected data, we used the bystander reaction dataset as input to a deep-learning model, BADNet, to predict failure occurrence. We tested different data labeling methods and learned how they affect model performance, achieving precisions above 90\%. 
We discuss strategies to model bystander reactions and predict failure and how this approach can be used in real-world robotic deployments to detect errors and improve robot performance. As part of this work, we also contribute with the ``Bystander Affect Detection" (BAD) dataset of bystander reactions, supporting the development of better prediction models.
\end{abstract}

\maketitle

\section{Introduction}

\begin{figure}[t]
\centering
\includegraphics[width=0.7\columnwidth]{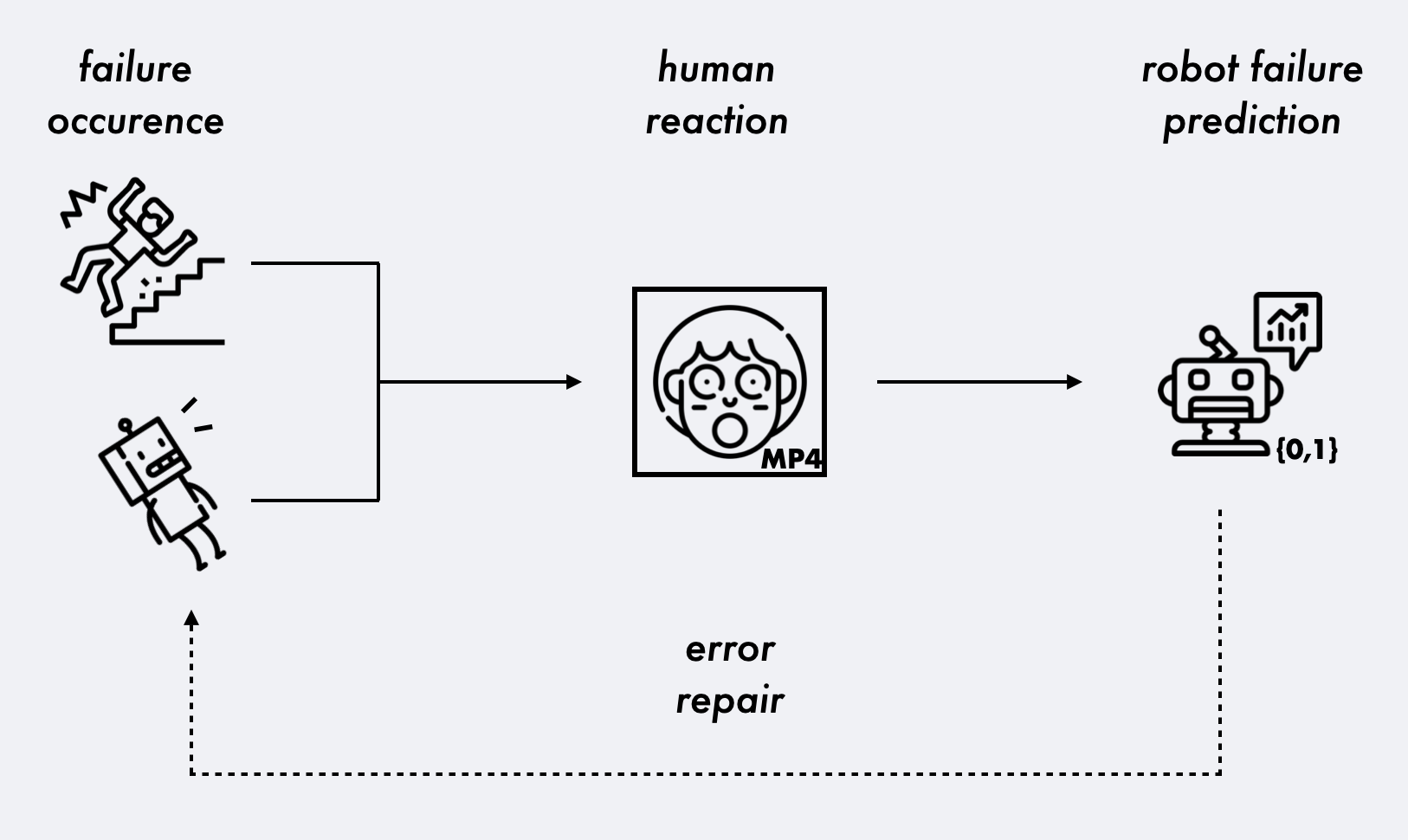}
\caption{Schematic overview of interaction intelligence through bystander response. A failure occurs (left), leading to a human reaction (center), which is used as data input for failure prediction from the robot (right). Adapted from \citet{bremers2023review}.}
\label{fig_schematicconcept}
\end{figure}

As robots get used in a wider array of tasks, more opportunities for robot failures arise. Increasing robots' awareness of their errors -- which we define as actions occurring ``not as planned" \cite{hollnagel1991phenotype} -- is a promising way to improve human-robot interactions \cite{bremers2023review,kontogiorgos2021systematic}. Prior work has studied how humans react to robot failures \cite{cuadra2021look,2020stiber,kontogiorgos2020behavioural}, revealing complex and multimodal responses. More recently, research in the field of Human-Robot Interaction (HRI) is investigating the potential of these social cues to inform robots that an error has occurred \cite{2022stiber,stiber2023using}. 
However, these works study settings where the human directly interacts with the robot under very controlled conditions that frequently involve explicitly designed failures performed by a limited number of robots. This might affect not only how naturalistic the elicited reactions are but also the potential for application in real-life scenarios in less controlled environments. As the integration of robots in human spaces creates \textit{bystanders} -- people not directly or deliberately seeking to interact with the robot, but nevertheless co-present with the robot -- we investigate the potential of using bystander facial reactions to develop robot failure detection models that might work across a broader variety of situations and which might more easily scale (first described by \citet{bremers2023review}, \autoref{fig_schematicconcept}). 

To achieve this, we designed an online data collection method that used webcam recordings to collect the Bystander Affect Detection (BAD) dataset\footnote{\url{https://bad-dataset.tech.cornell.edu/}} of \textit{human reactions} to failure. We release this dataset in an effort to enable others to further advance failure detection models, motivated by examples such as the ``Oops!" dataset \cite{epstein2020oops}, a compilation of short videos of moments of ``unintentional activity" which was made publicly available and led to significant developments towards models for failure prediction, intention inference, and problem localization \cite{epstein2021learning,das2021plsm,zhou2021temporal, xu2022probabilistic}. Finally, we show the potential of the BAD dataset through BADNet, a deep-learning model which achieves high performances in detecting human reactions to failure.
\section{Background and Related Work}

Detecting the occurrence of a physical task failure from a human bystander reaction builds on knowledge of human reactions to failure, as well as HRI-specific insights around failure detection. 

\subsection{Social Cues in Reactions to Failure}

People generate a wide variety of social cues in response to events, including acts of failure and success. Often, these social cues can communicate one's internal states with respect to the surrounding context. The range of social responses to error is widely studied in the behavioral science literature: facial expressions \cite{jabon2009automatically,lewis2010cultural,2022stiber}, physiological responses \cite{shirai2017sadness,yazmir2017act}, gestures and vocal expressions \cite{rossi2022generating,2020stiber}, among others \cite{kontogiorgos2021systematic,bremers2023review}. 

These complex social signals are easily understood by other humans but are challenging to implement as inputs for intention detection and recognition in robotic systems. 

\subsection{Leveraging social cues in HRI}
Recent work has investigated the potential of harnessing human social cues in response to robotic error. \citet{kontogiorgos2021systematic} use multimodal (acoustic and facial features) human reaction data for social error detection, specifically robot conversational errors. While their binary failure detection model achieves good performance (up to 77\% accuracy), the authors report the difficulties of transferability of failure detection models into new contexts, as well as in detecting different failure types. More recently, \citet{stiber2023using} leveraged facial expressions of humans interacting with a robot for error detection models across three HRI tasks: collaborative assembly, collaborative cooking, and programming by demonstration. Their work reinforces the idea that human reactions to failure are complex and vary according to the context of the failure, but facial expressions show promise as inputs for error detection models. \citet{bremers2023review} provide an overview of the existing literature in this space and discuss important opportunity areas for practitioners in HRI, including the need for datasets that cover a wider range of failure types and bystander responses. 

\subsection{Relevant datasets} 
We note that in most existing literature (as reviewed by \cite{kontogiorgos2021systematic, bremers2023review}), research studies were \textit{not} accompanied by the publication or release of raw datasets of people's responses to robot errors (\citet{stiber2023using}'s groundbreaking study featured a dataset of facial reactions to robotic errors, but includes only the facial action units associated with user reactions). This lack of available datasets is likely due to the fact that the studies to date feature data collected from \textit{in-situ} interactions; from a study perspective, this data has the most ecological validity, but the models trained on this data are not intended to work in any other context.

The machine learning community, in contrast, has made more use of larger-scale datasets that are published with the intent of enabling other researchers to develop general-purpose recognition models. \citet{epstein2020oops}'s ``Oops!" dataset, for example, contains a large video database of human-generated errors, which provides the computer vision community the opportunity to develop different models for action intention and failure recognition \cite{epstein2021learning}.

Similarly, the Affectiva-MIT Facial Expression Dataset (AM-FED), which features 252 reaction videos of people watching three popular 2011 Super Bowl commercials \cite{mcduff2013affectiva}, enabled other researchers to develop an LSTM algorithm for expressivity, which was posed as a possible metric for the engagement or entertainment quality of the stimuli \cite{srinivasa2017analysis}. This dataset was useful in part because it was collected ``in the wild" -- over the internet, with people's own webcams. This variability made the models trained on the AM-FED dataset more robust than models trained on other datasets of spontaneous expression in response to in-lab stimuli, such as the Belfast database \cite{sneddon2011belfast}.

\section{The Bystander Affect Detection (BAD) Dataset}
\label{sec:collection}

We introduce the Bystander Affect Detection dataset -- hereby termed the \textbf{BAD dataset} -- a dataset of videos of bystander reactions to videos of failures\footnote{\url{https://bad-dataset.tech.cornell.edu/}}. This dataset includes 2452 human reactions to failure, collected in contexts that approximate ``in-the-wild" data collection -- including natural variances in webcam quality, lighting, and background. 
Below, we describe the experimental design used for data collection and provide details on the dataset.

\subsection{Data Collection}

The BAD dataset was collected through an online survey. Participants were shown short videos of human- or robot-related failures while their reaction was recorded through the webcam of their computer. The \textit{stimulus} data included 40 failure videos, and 6 control (no error) videos. 

\subsubsection{Stimulus Videos}

The stimulus data consisted of videos sourced via Google and YouTube. Our search words included \textit{robot failure, human error, workspace failure, and professional tasks}. The resulting videos fell into the following 4 categories: human failures in professional settings, human failures in other settings, robot failure, and control video with the successful accomplishment of a task. We narrowed the selection down to 46 videos following requirements for videos to be non-violent, unstaged, uncaptioned, and having a clear failure. 

The resulting stimulus videos had a duration range of 6.0 to 35.8 s ($M\pm SD:14.3\pm6.6$ s).
Descriptive details of the stimulus video data can be found on the database website. 

\subsubsection{Data Recording}

\begin{figure}[t]
\centering
\includegraphics[width=0.7\columnwidth]{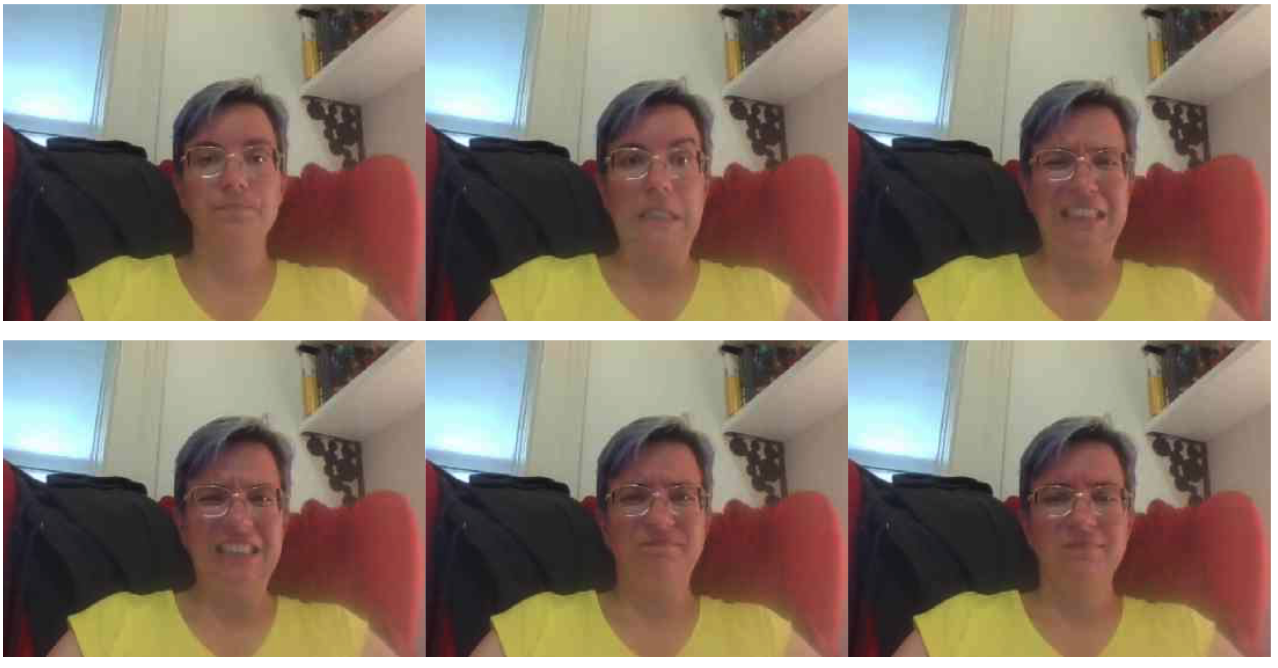}
\caption{\small Top to bottom, left to right: stills taken at 2-second intervals from a reaction video in the BAD dataset.}
\label{fig:responses}
\end{figure}

We conducted an online Qualtrics survey where participants watched stimulus videos while their laptop or computer webcam recorded their facial responses. Participants were not able to see their own image while the stimulus videos played. The order of stimulus videos was randomized. Due to the online and naturalistic setting of the experiment, the recorded reaction videos varied in terms of webcam quality, lighting, framing of the participant, and background. \autoref{fig:responses} shows an example reaction that was included in the BAD dataset.

Participants were recruited through Prolific. After giving informed consent and passing an attention check, participants were prompted to watch one video at a time while their reaction was being recorded. After each video, participants were asked to \textit{``Please describe in a few words what you saw in the video,"} and to indicate their agreement with the statements, on a 7-point Likert scale: \textit{``This context unfolded as expected in this video"} and \textit{``The task was completed successfully in the video."} These last prompts were targeted at understanding whether the participants perceived the video scene to depict a failure or not and the answers to these prompts are included with the BAD Dataset.

We included demographic information that participants provided to Prolific (i.e., ethnicity, country of birth, country of residence, nationality, language, student status, employment status) and additionally collected their age, highest level of education, and gender. Participants who watched at least 23 videos were compensated at the Prolific rate of \$12 per hour. Most participants took about 60 minutes to complete the survey. 

\subsubsection{Participants}

We recruited 109 participants through Prolific. Of these, we excluded 55 participants for one of two reasons: (1) failure to react to a minimum number of 30 failure videos (e.g., participant quit the survey), (2) technical difficulties in recording participants' reactions to the videos (insufficient video quality, insufficient lighting conditions, or incorrect body positioning). Therefore, the final dataset includes 54 participants' data. Two participants did not provide the demographic data via their Prolific profile. For the remaining 52 participants, ages range from 18 to 63 years old ($M\pm SD:27.8\pm8.8$). For gender identity, 24 participants self-identified as female and 28 as male. All participants have an education level of a high school diploma or higher. Ethnicities included white/Caucasian/European ($N=32$), Black/African ($N=9$), Mixed ($N=5$), and Other ($N=3$). The participants' most common countries of residence were South Africa, Portugal, Poland, Mexico, and Italy (all with at least 6 participants).

\subsection{Dataset Characteristics and Observations}

\subsubsection{BAD dataset}

The BAD dataset is made up of 2452 human reactions to 46 stimulus videos ($M\pm SD:45.41\pm1.4$ reaction videos per participant, max. $46$, min. $39$). Reactions vary in duration ($M\pm SD: 18.5\pm12.3$ s). If participants reacted multiple times to the same video (because of replaying the video), only the first reaction was kept. One example of a reaction video included in the dataset can be seen in \autoref{fig:responses}.

\subsubsection{High-Level Observations} 
\label{sec:observations}

To provide a brief overview of the dataset, we highlight observed phenomena in the participants' reactions to the different types of failure videos. These are based on general exploratory interaction analysis observations while handling the data. One or multiple of these events and aspects can occur simultaneously:

\begin{itemize}[noitemsep]
\item \textbf{Multimodality:} Reactions are complex and multimodal -- facial expression changes, sometimes accompanied by nodding or head shaking, changes in body position, or vocal outputs.
\item \textbf{Humor:} What makes a mistake funny? Some participants react with smiles and laughter, whereas others more reacting with worry and concern.
\item \textbf{Inter- and intra-participant diversity:} For the same video stimulus, reactions between participants differ in terms of the type of reaction and its magnitude. The same participant also displays diversity in their reactions across different video stimuli.
\item \textbf{Anticipation:} In cases where an error is predictable (e.g., a robot shakes back and forth before falling), human reactions can be spotted before the failure event occurs, and there is a slow buildup of reaction. This is consistent with prior observations \cite{2020stiber}.
\item \textbf{Non-reactions:} Some people never react at all -- maybe they were not even watching. These are ``false negatives", where detecting failure through bystander reaction would entail a challenge to any machine learning algorithm. In our data collection process, we built in steps (such as asking for a description of each video, and including an attention checker question) to control for participants' attentiveness to the video stimulus they observed, to reduce the effects of inattentive participant reactions on model training.
\end{itemize}

Other observations can be drawn from the analysis of participants' reactions and insights from behavioral science:

\begin{itemize}
\item \textbf{Empathy:} In some cases, the participant's facial expression seems to almost communicate a form of empathy, or correlate with their degree of relatedness to the robot or person undergoing the failure. This hypothesis is in line with the finding from \citet{kang2010your} that people's reaction to failure differs according to relatedness to the subject causing the failure. 
\item \textbf{Agency:} Related to empathy, there is variation in the apparent agency in the stimulus videos. For instance, a drone flying into a grill is assumed to be controlled by a human, whereas in the case of a robot falling over, it might not elicit the sense of human agency behind the error. This opens the question of whether human reactions are contingent on the perceived agency.
\item \textbf{Blame:} Similarly, sometimes the error comes across as accidental and unpredictable, other times it is the result of someone's actions and could have been prevented. These contrasts may partially explain the intra-participant variability in responses to observed errors.
\item \textbf{Humanity:} Another factor that may affect empathy, type, and magnitude of reaction is the robot's appearance. Anthropomorphic robots might invoke a greater response than non-anthropomorphic robots do \cite{2001bickmore, Deng_2019, kontogiorgos2020behavioural}. This too calls for further investigation.
\end{itemize}

\section{BADNet for Failure Detection}

To demonstrate a use case of the BAD dataset, we present BADNet -- a deep learning model to detect failure occurrence in the environment based on bystander human reaction (RBG video frames) as the single model input.

We formulate this problem as sequential decision-making. At any time step, $t$, the environment (visual data) is captured as a state variable $s_t \in S$. The model outputs a prediction $p_t \in P$. It chooses between a binary output: \textit{failure occurred} (1) or \textit{failure did not occur} (0). 

\subsection{Visual features and labeling}
\label{subsec:labeling}

\begin{figure}[t]
\makebox[0.7\textwidth][c]{\includegraphics[width=0.7\textwidth]{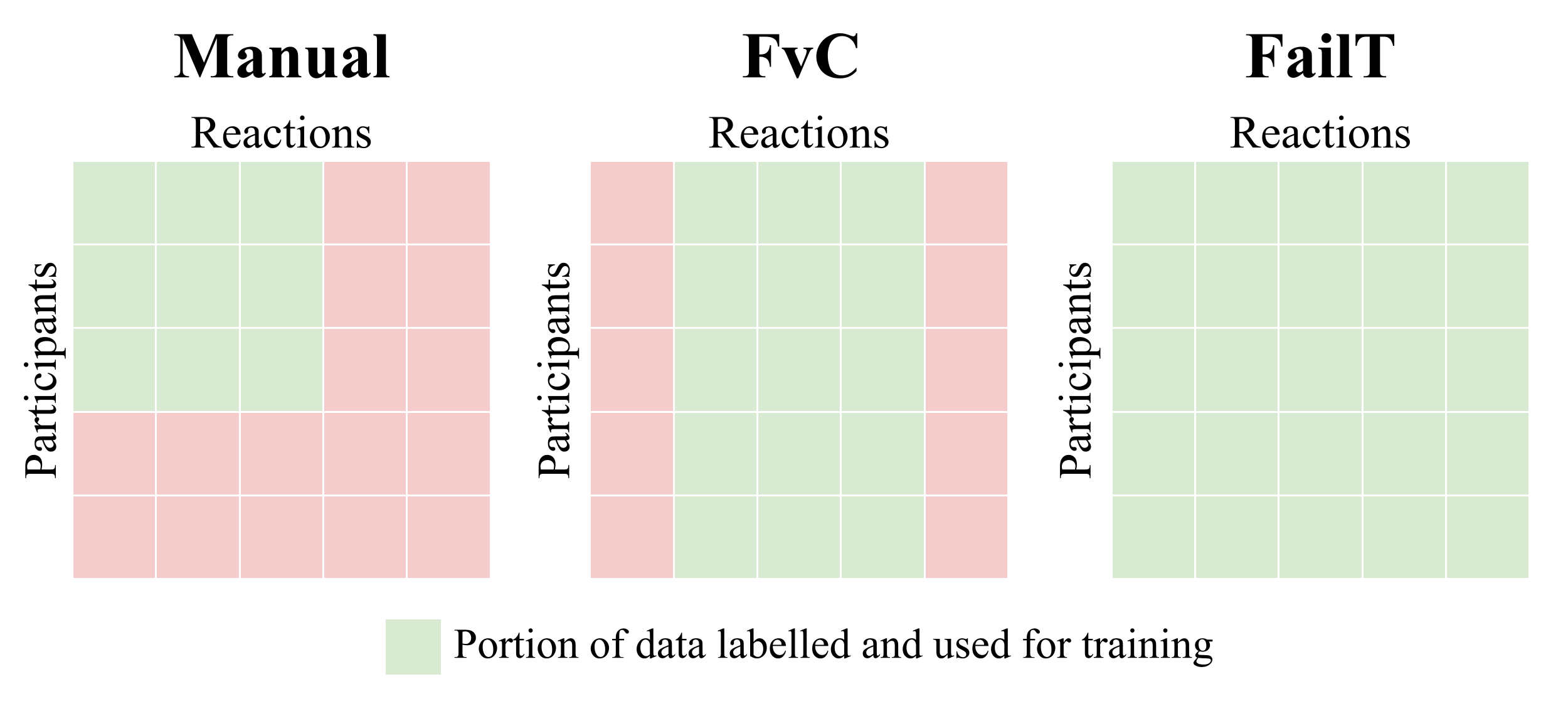}}
\caption{Visual representation of the different labeling methods and the corresponding portion of the total dataset used for training.}
\label{fig:labeling}
\end{figure}

The model input comprises single RGB image frames from the BAD dataset videos (see Section \ref{sec:collection}). We post-processed the video into static frames at a 30-frames-per-second rate. The frames are labeled into one of two categories, based on the state of the participant at that moment: \textit{reaction} (1) or \textit{neutral state} (0). Due to limitations in the data collection process, we took an exploratory approach and adopted three different methods for labeling the data, which we detail below and illustrated in \autoref{fig:labeling}.

\begin{description}[align=left, leftmargin=0em, labelsep=0.2em, font=\textbf, itemsep=0em,parsep=0.3em]

\item [Single-blind labeling (manual):]
In this approach, we naively labelled data according to \textit{when humans display a reaction}, independently of the time when the failure occurred (that is, anticipatory or delayed reactions are also labeled \textit{reaction to failure}). According to this criterion, two researchers labeled reaction videos from 12 participants chosen at random ($11.08\pm 1.97$ videos per participant, min. $10$). All labeled frames were extracted and used for training our models in a total of 37472 frames (24191 neutral states, 13281 reactions). 

\item[Failure-vs-control (FvC) labeling:]
Alternatively, we considered an automated labeling method where only reactions to a portion of the stimulus videos were included. The stimulus videos included 6 control videos without failure (Video IDS: $686, 683, 700, 707, 714, 721$, refer to database website). We treated participants' reactions to these videos as a ground-truth \textit{neutral state} and labeled \textit{every frame} accordingly (0). To balance the dataset, we selected 6 failure stimulus videos, where the failure occurs gradually (QIDS: $160, 178, 220, 406, 130, 609$). Analogously, we considered participants' reactions to these videos as a ground truth \textit{reaction to failure} and labeled every frame accordingly (1). We extracted a total of 31212  frames (14505 neutral states, 16707 reactions) across all 54 participants.

\item[Failure-time (FailT) labeling:]
Finally, we used an alternative automated labeling method based on the \textit{exact moment} when a failure occurs in the stimulus video. For each reaction video, we consider the time stamp when the failure occurred in the corresponding stimulus video. Afterward, we extract 60 randomly-selected frames from the moments \textit{before} failure, which is labeled as  \textit{neutral state} (0), and 60 frames from the moments \textit{after} failure, which are labeled as  \textit{reaction to failure} (1). 
We extracted a total of 262723 frames (135764 neutral states, 126959 reactions) across all 54 participants.
\end{description}

\subsection{Leveraging deep learning for failure detection - BADNet}

We trained three models with the same architectures, one for each labeling method applied (Section \ref{subsec:labeling}). Below, we describe the general architecture, the training procedure, and the performance of the models.

\subsubsection{BADNet Architecture}
We introduce BADNet, a CNN-based model for detecting failure through bystander reactions. The model architecture is fundamentally an adaptation of AlexNet, which has historically been one of the simplest state-of-the-art image classification approaches \cite{krizhevsky2017imagenet}. \autoref{t:architecture} displays the full architecture of the model. Optimization in training and regularization procedures borrow from implementations of naturalistic dataset domains that include environmental sounds \cite{ebrahimpour2020end} and brain signals during music listening \cite{ramirez2022image}. Specifically, this is implemented through Global Average Pooling (GAP) after the last convolutional layer, which greatly reduces the number of trainable parameters in the model by shrinking the dimensionality of the data. 

\begin{table}[th]
    \centering
    \caption{Details of the model architecture}
    
    Input: Raw RGB webcam images of participants, resized to 224x224 \\Output: \textit{failure occurred} (1) or \textit{failure did not occur} (0)
    \vspace{.5em}
    \scalebox{0.9}{
    \begin{tabular}{c|c c c}

    \hline
    Layer Type & Filter Size & Input & Output\\
    \hline
    \hline
    Conv2D & $4 \times 4$ & $224 \times 224 \times 3$ & $112 \times 112 \times 16$ \\
    BatchNorm2D & - & - & - \\
    \hline
    Conv2D & $4 \times 4$ & $112 \times 112 \times 16$ & $56 \times 56 \times 32$ \\
    BatchNorm2D & - & - & - \\
    \hline
    Conv2D & $4 \times 4$ & $56 \times 56 \times 32$ & $28 \times 28 \times 64$ \\  
    BatchNorm2D & - & - & - \\
    \hline
    GAP & $28 \times 28$ & $28 \times 28 \times 64$ & $1 \times 1 \times 64$ \\
    \hline
    FC1 & - & 64 & 128 \\
    BatchNorm1D & - & - & - \\
    FC2 & - & 128 & 2 \\
    \hline
    \end{tabular}}
    \label{t:architecture}
    
\end{table}

While not depending on feature extraction of the data, this model only has 50,674 trainable parameters, which roughly translates to 652 Kilobytes of data when the model is saved and exported. Other typical computer vision architectures, such as smaller versions of ResNet \cite{he2016deep}, are on the order of ~1 million trainable parameters.

\begin{table*}[ht]
\centering
\tabcolsep=0.11cm
\caption{BADNet model performance summary}
     Grand performance summary of all BADNet models per each labeling strategy of the dataset. \\ We report $M\pm SD$ across the 4 validation folds.\\
    \vspace{1em}
\begin{tabular}{c|c|c|c|c}
\hline
 \textbf{Models}& \textbf{Recall\%}& \textbf{Precision\%}& \textbf{F1\%}& \textbf{Kappa\%}\\[2pt]
 \hline
 \textbf{Manual} & $95.26 (\pm 1.60)$& $95.74 (\pm 1.34)$& $95.31 (\pm 1.57)$& $89.92 (\pm 3.36)$\\[2pt]
  \hline
\textbf{FvC} & $95.02 (\pm1.77)$& $95.12 (\pm 1.67)$& $95.00 (\pm1.78)$& $89.95 (\pm 3.57)$\\[2pt]
  \hline
\textbf{FailT} & $89.88 (\pm 0.486)$& $90.03 (\pm 0.468)$& $89.88 (\pm 0.481)$ & $79.77 (\pm 0.95)$\\[2pt]
  \hline
\end{tabular}
    \label{t:performance_summary}
\end{table*}

\subsubsection{Hyper-parameter tuning}
We took an end-to-end computer vision modeling approach. Such an approach aims to let the model learn how to extract features through its convolutional layers; therefore, filter size was kept at a $4 \times 4$ with a stride length of 2. Network weights were initialized using a He distribution, which serves as a type of random weight initialization that defines the model's priors to ease convergence when nonlinear transformations are applied, such as in our use of ReLu activation functions in all intermediate layers \cite{he2015delving}. Finally, regularization of gradient descent in the training procedures was focused on the addition of dropout between intermediate layers at a $[0.15,0.2,0.2,0.6]$ value order for the \textbf{Manual} data, $[0.1,0.15,0.15,0.6]$ for the \textbf{FvC} data, and $[0.025,0.025,0.05,0.35]$ for the \textbf{FailT} data.

\subsubsection{Evaluation Metrics}

Performance was evaluated based on the macro-average of the following measures: recall, precision, F1-score, and Cohen's Kappa. Recall refers to classification performance as a ratio of true positives over true positives and false negatives (probability of detecting failure if a failure occurred), while precision is the ratio of true positives over true positives and false positives (probability of a failure detection being true). F1 is the harmonic mean of both recall and precision, while Cohen's Kappa is a metric used as a proxy for data labeling noise \cite{cohenskappa}. This works by normalizing classification performance by the random chance of agreeability between existing data labels and predicted model labels, which makes it the strictest performance metric used here.  

\subsection{Training and Model Performance}

For training and evaluating model performance, we used a 4-fold cross-validation process (75/25\% train-test split). The training was thus generalized within participants, across time, and with randomly unbalanced examples per participant. A summary of the performance for each of the three models can be seen in \autoref{t:performance_summary}.

Additionally, we trained and tested the models for generalization to new unseen participants. This was performed through a single-randomized-participant hold-out for the test set, repeated 4 times with different participants.
Performance for the \textbf{FailT} and \textbf{Manual} models did not go above random chance (50\%). Interestingly, the \textbf{FvC} model showed an average performance of 63.65\% across the participant hold-outs, reaching values as high as 85.70\%.

\section{Discussion}
\label{sec:discussion}

The use of bystander human behavior as an input in automated robotic systems provides an efficient way to expand robots' perception of their environment. This, in turn, can be applied to expand robots' task-related skills, such as reactive performance in task completion, as bystander reactions are used as ``sensors" for detecting how the task is being carried out. 

\subsection{The BAD dataset}
One goal with the BAD dataset was to design the data collection method so that the approach was scalable, thus more likely to generate bystander affect detection that is robust and generalizable.
The use of an online survey not only allows for a practical collection of data that is diverse in demographic features but also mimics an ``in-the-wild" interaction. This approach has limitations (see Section \ref{subsec:limitations}), but also provides a more realistic insight into how humans react to failure when they are not directly engaged in the ongoing interaction. By including manipulation control questions, users of the dataset can discern when a lack of identifiable reactions is due to inattention or due to natural human response.

We also included high-level observation of some of the reactions to failure present in our dataset. The diverse set of phenomena reported is proof of the complexity and richness of human reactions, even when the subjects are bystanders and not direct actors in a setting.
The BAD dataset is made available for other researchers, with the intention of improving recognition of and response to bystander behavior in robotic systems, which we believe will  improve human-robot interaction and robot task performance.

\subsection{BADNet for bystander affect detection}

We introduced BADNet, a deep learning model that successfully identifies failure through bystander reactions. This unimodal model leverages raw RGB image inputs (single frames from human reaction videos) and achieves strong performance, regardless of the strategy used for labeling data. We provide three strategies for the data labeling process (\textit{Manual, FvC} or \textit{FailT}). These stand on different assumptions made about the data: the length of human reaction to failure; whether anticipation of failure is accounted for in the detection; or the ultimate goal of the detection (e.g., know \textit{if} a failure has occurred versus \textit{when} a failure has occurred). 

Different labeling methods lead to different model performances. Both the \textbf{Manual} and the \textbf{FvC} models achieve high performance, with Cohen's Kappa indicating good agreeability across the data. While performance in the \textbf{FailT} model is lower, the standard deviation is also the lowest. We highlight the standard deviation reported for Cohen's Kappa, which is below 1\% despite the model being trained in the largest dataset of the three. Lower and more inconsistent results are seen for the leave-one-participant-out approach. This means that a period of training might be necessary for adaptation to new participants outside this dataset. We also note that, given that this is an ``in-the-wild" data collection, heterogeneity within the data is to be expected. Like in many machine learning applications, it is expected that an increase in participants will yield better generalization to new people. Generalization across time and with randomly unbalanced examples per participant holds strong, given the amount of data provided per participant. 
Further exploration of how the dataset's feature dimensions generalize is needed, in order to understand how we can better adjust models for practical use.

Balancing the model size and performance trade-off is vital in highlighting what is possible with the dataset. BADNet is a ``lightweight" model which works on raw data. This opens up opportunities for real-life deployment. Given that the exported model is small enough to fit in a floppy disk, the model might be suitable for implementation in hardware and data-storage solutions that are limited in bandwidth and memory. 

\subsection{Limitations}
\label{subsec:limitations}

In collecting this dataset, the technical and experimental issues that arose were informative of the challenges in collecting online ``in-the-wild" video data, namely the aspects of image quality (webcam resolution, lighting conditions), participant frame placement (not centered on the frame), data recording and labeling (time lags due to internet connection speeds) and participant behavior (distractions from the environment); nonetheless, the real-life deployment of these systems will be bounded by the same challenges. This means that models and analyses that stem from the BADNet dataset are more robust to extrapolation into real-use cases, especially if developed models follow an end-to-end processing approach and demonstrate strong performance with a relatively small amount of trainable parameters, such as is the case with BADNet.

\begin{figure}[t]
\centering
\includegraphics[width=0.4\textwidth]{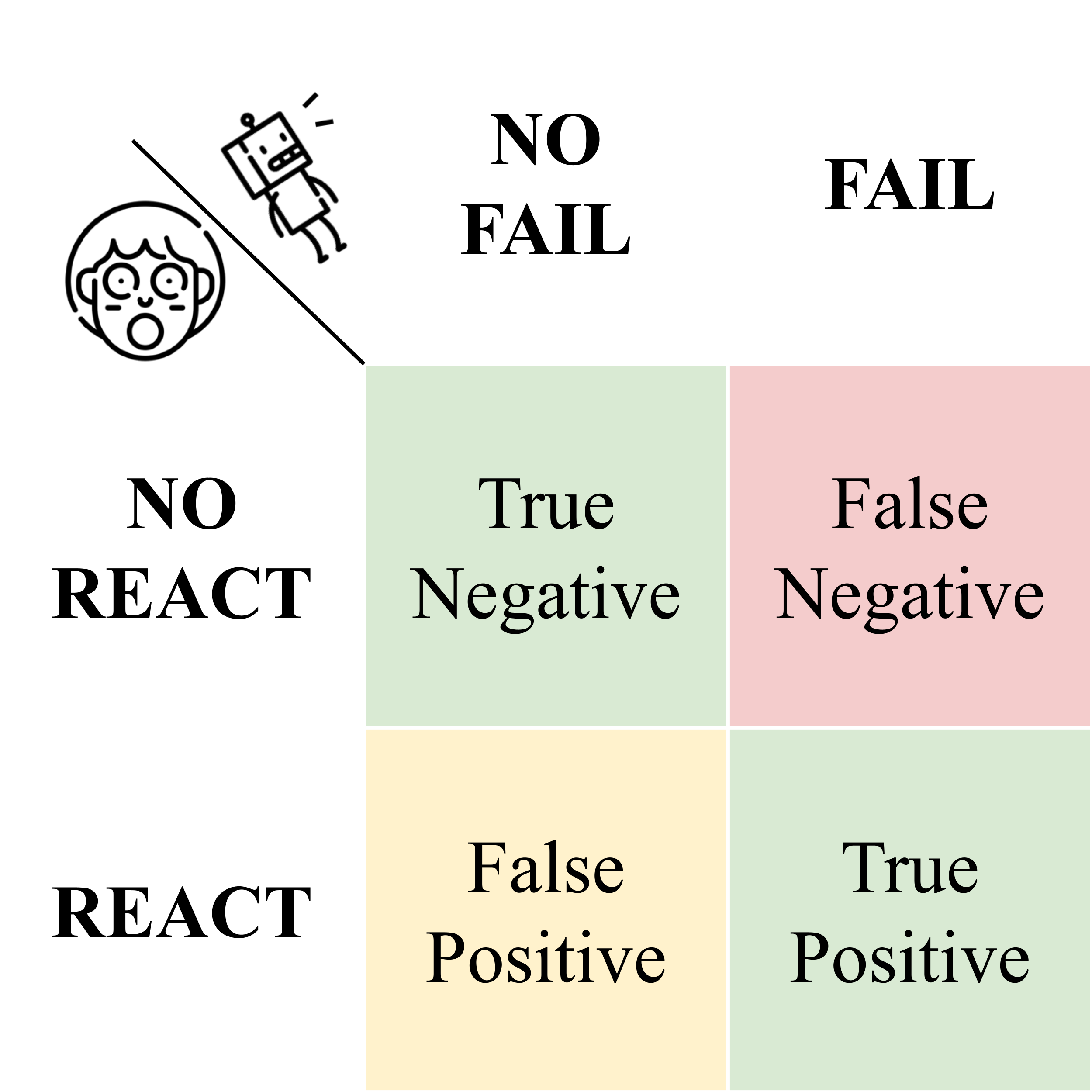}
\caption{Detection cases for BADNet. The ground truth is given by whether failure occurred (FAIL) or not (NO FAIL), but the model inputs are bystander reactions.\protect\footnotemark}
\label{fig:falseneg}
\end{figure}

We must also recognize a limitation of the model we propose as it is not optimized to detect false negatives (see \autoref{fig:falseneg}). Given that the human reaction is the input of the model, failure can only be detected when humans react to it or when we already know that failure is occurring. 
In addition, ideally, a model should be able to distinguish spurious reactions from reactions to failure (false positives). Future work is needed to understand the generalizability of specific behaviors and the possibility of models developed to detect a taxonomy of behaviors. For this, a set of negative examples (when humans react, but not in response to failure) would have to be provided and optimized for.

Our dataset, which is sampled through Prolific, also shows a limitation in its demographic variance, specifically with skews in ethnicity, location, and age. Researchers developing models based on the dataset are advised to keep these skews in mind, depending on the model application.  

Finally, we would like to point out to future users of the BAD dataset to keep in mind ethical considerations while using the non-anonymized reaction videos dataset for developing systems. As part of this consideration, the non-anonymized dataset will not be freely downloadable but access can be requested along with the presentation of a research protocol or data use agreement that protects participants.  

\footnotetext{Icons by Wanicon and Freepik via www.flaticon.com.}

\subsection{Future Work}

BADNet proposes one approach for leveraging bystander reactions to inform robot task performance and failure. 
Future work could involve expanding BADNet by including multimodal sources of data. For human social cues, one could include vector representations of features such as facial expressions \cite{stiber2023using,2013cid}, prosodic/audio data \cite{goedicke2021acoustically}, and body position or body motion \cite{2017trung}. Contextual data inputs may also be used, such as additional sensor data from the environment, to allow the robot to use information about its surroundings to detect what or when something failed. 
Additional information streams also lend themselves to exploring alternative classifier architectures and techniques, namely the use of recurring neural networks (RNNs) or Reinforcement Learning (RL), whose potential remains mostly unexplored in this field. These methods may be contingent on the expected use case and goals -- is the system personalized for one user, or generalizable to bystanders that the system has never seen before? This will impact how the model should be trained and how performance is defined.

Finally, this dataset can be expanded to allow for augmenting our understanding of how bystanders react to failure. As suggested by \citet{bremers2023review}, one can explore if bystander reactions to humans or robots' failing participants are related to empathy and social identity \cite{kang2010your}. In-person versus online reactions are also likely to be different, so future studies with online failure detection are needed to validate the transferability of the model. Finally, we note the need to account for demographic and cultural factors which can also play a role on the reactions elicited and performance of the system \cite{2014salem,2015salem}.

\section{Conclusion}

This work contributes with the BAD dataset and BADNet, important tools to advance the field of human social cue recognition for failure detection, that extend on existing work by including a wide variety of failure types and collecting bystander reactions in uncontrolled environments, through an online data collection method. This dataset can be used to develop models that detect when something goes wrong -- an important step in building trustworthy and seamless human-robot interactions. We also provide such a model, BADNet, which detects failure with precisions up to 95\% and is ``light-weight", thus suited for real-life and real-time applications. We invite researchers to use these tools and integrate them into human-robot interaction loops.

\section*{Acknowledgements}
This data was collected under exempt IRB protocol \#1609006604 of Cornell University with informed consent of participants. The authors would like to thank members and collaborators of the Cornell Tech research community for their help and feedback, and in particular Ilan Mandel, Rei Lee and J.D. Zamfirescu-Pereira, for their contributions to earlier versions of this project.

\small
\balance
\bibliographystyle{IEEEtranN}
\bibliography{bibliography.bib}

\end{document}